**Title:** A Finite-Time Technological Singularity Model With Artificial Intelligence Self-Improvement

**Running title:** A Finite-Time Singularity Model

**The full name of the author:** Ihor Kendiukhov

**Affiliation:** Faculty of Economics and Business Administration, Humboldt University of Berlin, Unter den Linden 6, 10099 Berlin, Germany

**e-mail:** kenduhov.ig@gmail.com

# A Finite-Time Technological Singularity Model With Artificial Intelligence Self-Improvement

**Abstract**

Recent advances in the development of artificial intelligence, technological progress acceleration, long-term trends of macroeconomic dynamics increase the relevance of technological singularity hypothesis. In this paper, we build a model of finite-time technological singularity assuming that artificial intelligence will replace humans for artificial intelligence engineers after some point in time when it is developed enough. This model implies the following: let A be the level of development of artificial intelligence. Then, the moment of technological singularity n is defined as $\lim_{t \to n} A(t) = \infty$. Thus, it happens in finite time. Although infinite level of development of artificial intelligence cannot be reached practically, this approximation is useful for several reasons, firstly because it allows modeling a phase transition or a change of regime. In the model, intelligence growth function appears to be hyperbolic function under relatively broad conditions which we list and compare. Subsequently, we also add a stochastic term (Brownian motion) to the model and investigate the changes in its behavior. The results can be applied for the modeling of dynamics of various processes characterized by multiplicative growth.

1. **Introduction**

Recent advances in the development of artificial intelligence, technological progress acceleration, long-term trends of macroeconomic dynamics increase the relevance of technological singularity hypothesis and make severe economical and societal changes connected with technological and scientific advancements more possible. For the overview and discussion regarding technological trends in the context of singularity, see Kurzweil (2005), Schmidt and Cohen (2013).

Various organizations, professionals, and researches related to the AI industry (Kurzweil, 2010, Sandberg, 2010) claimed that some form of technological singularity may be reached in 21 century, although they define singularity very differently. Even if the event remains to be extremely unlikely, its huge potential impact justifies a more detailed investigation of the problem.

Mathematical models of technological singularity are widely used to predict and estimate important events in the history of humanity, phase transitions, and technological breakthroughs, as well as long-term growth rates of economies, populations, and technologies. Most of such models imply hyperexponential growth rates of variables studied.

Finite-time technological singularity concept was proposed in various models:

$$\lim_{t \to n} A(t) = \infty \qquad (1)$$

where *A* is the level of development, which can be expressed as GDP, technology, intelligence, progress, etc., and *t* is time.

Although an infinite level of development of artificial intelligence cannot be reached practically, this approximation is useful for several reasons. Firstly, vertical asymptote is a way to model a phase transition, a change of regime, a disaster, or a breakthrough. Although the "infinite" level of development cannot effectively be reached, the model that describes it can thus predict the time of important shifts in the growth trends. Secondly, even if the model with finite-time singularity may be not relevant on the whole time interval from zero to infinity, it can describe growth rates precisely for a limited interval, as it was with the demographic growth model by Kapitza (1996). Thirdly, the models allow describing several growth regimes with extreme acceleration, which is not possible without finite-time singularity as a consequence. Fourthly, the definition of technological singularity as a finite-time infinite limit of a grows function is more mathematically and analytically univocal than alternative definitions.

Although it is definitely not clear whether technological singularity will or may be reached or not, possible payoffs and risks are extremely high. Furthermore, even if the singularity will not be reached, development models can be helpful in the study of various local processes in the society and the economy with accelerating growth dynamics.

Of course, explosive growth resulting in the technological singularity (whatever is meant by it) must not necessarily be hyperbolic or equivalent. For example, if the singularity is defined as the moment when the development of artificial intelligence exceeds the development of human intelligence, preceding growth dynamics may be linear or even sublinear.

The paper is structured as follows. Previous studies overview is introduced in section 2. The results of the study (models and empirical estimations) are described in section 3. In subsection 3.1, weak and strong forms of technological singularity are defined and discussed. In subsection 3.2, the basic model with a linear connection between AI growth rate and engineer intelligence is introduced and evaluated. A more general version of the model is given in subsection 3.3. The impact of changes in the underlying assumption on the estimated time of singularity is demonstrated here. In subsection 3.4, the models are extended by adding stochastic terms to them. In subsection 3.5, we explore some other ways to reach finite-time singularity even if with weaker assumptions on the AI self-improvement. In subsection 3.6, the reverse task is solved. Namely, conditions leading growth dynamics function to finite-time singularity are defined. Discussion is given in section 4, and conclusions are presented in section 5.

2. **Previous Studies**

Since the term "singularity" was popularized by Vernor Vinge (1993) in his essay "The Coming Technological Singularity", various analytical models of singularity based on different definitions were introduced. For a comprehensive overview of these models, we refer to Sandberg (2010). For the purposes of our analysis, we are interested in the overview of finite-time singularity models.

Basically, the need to use finite-time models can be derived from these words of Johansen and Sornette (2001):

"Singularities are always mathematical idealizations of natural phenomena: they are not present in reality but foreshadow an important transition or change of regime. In the present context, they must be interpreted as a kind of "critical point" signaling a fundamental and abrupt change of regime similar to what occurs in phase transitions."

The idea of "intelligence explosion" was mentioned by Good (1965), although finite-time singularity analytical models were not developed explicitly. Similar ideas in verbal form were expressed by various authors in recent decades (Eden et al., 2012).

As an example of a simple early finite-time singularity model, we can take the model of Solomonoff (1985). It described the growth of an effective computer science community (consisting of humans and AI) as a function of the rate of money spent on AI and the amount of computing power per dollar, which itself grows exponentially. Moravec (1999) builds another simple finite-time singularity model, where the growth trend at the beginning of the process is similar to exponential, although he states that the assumptions leading to finite-time singularity are too optimistic.

A little bit more advanced models, such as city economics (West et al., 2007), deliver finite-time solutions as a special case.

Attempts to investigate the concept of technological singularity in the context of the long-term macroeconomic growth model also lead under some assumptions to finite-time growth explosion. The model of Kremer (1993) has this feature. The model states that the interplay of means of production can create finite-time singularity even if each factor, taken separately, cannot. Another model connecting population and technology growth is created by Taagepera (1979). Under some parameter values, the growth in the model is becoming hyperbolic. Hanson (1998) essentially extends the scope of tools of economics which can be applied for the analysis of technological singularity resulting in fast intelligence explosion. A similar effect can be achieved via a broad implementation of brain emulation technologies (Hanson, 2008).

Romer (1986, 1990) explores the possibility of the growth rate of economic output tending to infinity. Generally, his models of endogenous growth can be used as a starting point of various modifications leading to finite-time singularity.

However, models of explosive technological growth were not popular among economists, partially because of the lack of empirical evidence, partially because they required the lack of diminishing returns in production functions (Jones, 1995).

In recent times, the concept of singularity received even more attention from economists. Nordhaus (2015) builds various economical empirical tests of the singularity hypothesis, although not making distinctions between infinite-time and finite-time singularity. The results of these diagnostic tests tend to show that the singularity is not near, yet no specific statements in the form of analytical models describing the specifics of the singularity, if it happens, are given. Generally, the singularity is considered here as a significant acceleration of growth. Nordhaus also states that "there is remarkably little writing on Singularity in the modern macroeconomic literature".

More optimistic, although less quantitative predictions and empirical observations are made by economists Brynjolfsson and McAfee (2014).

In this study, we propose a model that described solely the dynamics of artificial intelligence, without referring to the economic means of productions, demographic parameters, or intersectoral relationships. Thus, we show a scenario of how finite-time singularity can be achieved just via the internal dynamics of intelligence development. In this case, regardless of the nature of the relationship between the intelligence and various macroeconomic and technological variables, the impact of the intelligence development approaching infinity on all variables will obviously be huge. In other words, the near-infinite level of development of artificial intelligence would imply necessarily dramatic changes in the development of technology and economy, with resulting phase transition making further forecasts impossible. In other words, rather than predicting specific numerical values of various indicators at the moment of singularity, the model attempts to predict the time of singularity whereas the singularity itself is characterized rather qualitatively.

Empirically, both economic growth and artificial intelligence growth appear to be exponential at the moment (Easterly, Levine, 2001, Jones, 2016, Brock, Moore, 2006). However, the population followed a hyperbolic trend before 1970 when the inflection point occurred (Korotayev, 2007, Golosovsky, 2010). Current exponential growth trends are fully consistent with the proposed model.

### 3. Results

#### 3.1. Definitions

There are multiple definitions of technological singularity. Most of them are qualitative: they refer to a period or a moment when some explosive and extraordinary positive (in terms of the rate) event happens to technological growth. Sometimes it means that AI exceeds the level of human or intelligence, in other cases, it means the fact of irreversibility or uncontrollability of technological progress. Mostly the question of technological singularity is studied in the context of the development of artificial intelligence. It is supposed that the growth of artificial intelligence will reach a critical point at some time, and this critical point is defined as the singularity.

There is always an essential level of ambiguity in these explanations. Since we can take different measures of machine intelligence and compare it with human intelligence differently, model predictions may vary a lot. Even if the machine intelligence growth trend is robust and known, it may deliver a clear understanding of when the singularity may come only in rare cases. For example, let us consider the case of the exponential growth of artificial intelligence. Although it is clear that the rate of this growth exceeds (or will exceed sooner or later) the growth rate of human intelligence (which is put to zero in most of the similar models) it tells close to nothing either about the time of the singularity or about the possibility of the singularity itself. Indeed, if the difference between human intelligence and the current level of artificial intelligence is large enough, the singularity may not happen at any reasonable point in the future, even if the growth is exponential. It the ratio of human intelligence to artificial intelligence approaches infinity, then the singularity will occur infinitely late.

Hence, in order to avoid the ambiguity resulting from the vagueness of the definition of intelligence and its estimation, let us differentiate between weak and strong definitions of singularity.

*Weak singularity*: machine intelligence exceeds human intelligence.

*Strong singularity*: machine intelligence approaches infinity in finite time.

Actually, the strong definition of singularity is anticipated by these words of Good (1965):

"Let an ultraintelligent machine be defined as a machine that can far surpass all the intellectual activities of any man however clever. Since the design of machines is one of these intellectual activities, an ultraintelligent machine could design even better machines; there would then unquestionably be an "intelligence explosion", and the intelligence of man would be left far behind. Thus the first ultraintelligent machine is the last invention that man need ever make."

If we apply the strong definition of singularity, we are likely to avoid the error of mistaking non-singularity for singularity. In other words, *if the strong singularity is achieved, then any other singularity is achieved as well.*

By implementing the definition of strong singularity, we get rid of ambiguity. If we cannot agree on specific criteria of achievement of singularity under various models (which function value means singularity), we can use the strong definition of singularity, for if $A \to \infty$, then any value is achieved.

Singularity here is defined further in the paper as a moment in *n* time when *A* in formula (1) approaches infinity. This type of singularity is known in the literature as a finite-time singularity (Sandberg, 2010). We shall later denote the moment of singularity calculated as the distance from starting time in the model in years as $t_s$.

Of course, artificial intelligence neither can reach the "infinite" value of development, nor it can grow nonstop as fast as it can be allowed mathematically since there are physical burdens of AI development (Lloyd, 2000). Yet these burdens are extremely high and, considering our level of intelligence, we can ignore them. Our goal here is just to show what long-term trends in AI development can be and what explosive scenarios are possible. Undoubtedly, different types of limitations can change and stop these trends, but first, we need to investigate basic options.

### 3.2. Basic model

Let us suppose that the growth of artificial intelligence is exponential and depends on some constant coefficient $k$ and the intelligence level of the engineer $I$. In the current case, the engineer is, predictably, a human. Thus, $I$ is the human intelligence level. Hence,

$$dA = kIAdt \qquad (2)$$

Then, the solution of the equation (2) is:

$$A = ce^{kIt_1} \qquad (3)$$

There are a lot of measures of "intelligence". However, most of them are qualitative and thus cannot be used in our analysis. Quantitative measures are, on the other hand, not very representative, since they are just some correlates of what we usually define by "intelligence". But the non-perfect correlation of these quantitative measures with "actual intelligence" is less problematic in the case of finite-time singularity models. All the implications can be applied for different measures and further development of the measure of intelligence can make the models more precise and powerful.

We can take processor performance as a measure of artificial intelligence development. Empirical data (Denning, 2017) shows that it doubles every 18 months (1,5 years). Since we are interested in relative measures, we suppose that for $t = 0$, $A = 1$. Then, $I$ is the value that shows how many times human intellect at the moment exceeds the artificial one.

$$\begin{cases} R^{t_1} = ce^{kIt_1} \\ ce^0 = 1 \end{cases} \qquad (4)$$

where $R$ is annual growth rate.

Then,

$$c = 1; k = \frac{ln 1,5872}{I} \qquad (5)$$

An illustration of exponential growth dynamics is given in figure 1.

But what is the value of $I$? If we compare directly computation performance measures in FLOPS, the results are the following. Estimates for brain performance in FLOPS vary by orders of magnitude, from $3 * 10^{13}$ to $10^{25}$ FLOPS. The median estimate is $10^{19}$ FLOPS (Sandberg, 2016). Modern computers (such as Summit) can reach the speed of $10^{17}$ FLOPS. Based on this, the difference between human and modern artificial intelligence measured in this crude way is several orders of magnitude, with the median estimate of 2 orders. Then, estimation for $k$ can be, for example,

$$k = \frac{ln 1,5872}{100} \approx 0,00462 \tag{6}$$

But the growth dynamics changes when $A$ reaches the level of $I$. From now on, it is more efficient to apply artificial intelligence for the design of artificial intelligence. AI becomes an engineer. Analytically, it can be written as

$$dA = kA^2 dt \tag{7}$$

Then,

$$A = \frac{-1}{kt_2 + c} \tag{8}$$

But in this model for $t_2 = 0$, $A = I$, then

$$I = -\frac{1}{c} \tag{9}$$

and

$$A = -\frac{1}{kt_2 - \frac{1}{I}} \tag{10}$$

Is it reasonable? The formulas support the intuitive conclusion: higher growth rate and human intelligence today mean higher speed of reaching the singularity.

An illustration of hyperbolic growth dynamics is given in figure 2.

Hence, a positive deviation from the linear trend on a log scale can be a sign of the approaching of finite-time singularity.

If $I=100$, then $A = I$ when $t_1 \approx 10$:

$$t_1 = \frac{\ln 100}{\ln 1,5872} \approx 9.97 \qquad (11)$$

Taking into account formula (10) and based on the definition in formula (1), the moment of singularity is:

$$t_2 = \frac{1}{r} \qquad (12)$$

where $r = \ln R$

For our $r$, $t_2 \approx 2,16$.

Hence, the moment of singularity is defined only by the initial growth rate of artificial intelligence (in our case it is processor performance) and by the level of human intelligence relative to the level of artificial intelligence.

Combining $t_1$ and $t_2$, we get that for the parameters mentioned above, time distance of singularity $t_s$ is approximately equal to 12 years.

In general, the hyperbolic growth can begin earlier than AI reaches the level of human intelligence. It may happen because AI, still being less generally intelligent than humans, will be more proficient in the development of AI or in some corresponding areas, such as the production and development of computers. Moreover, the emergent effects of cooperation between humans and AIs may lead to a hyperbolic explosion (or similar trends). Evidence suggests that in some tasks (like chess) when cooperating with AI, humans are able to outperform both humans and AIs operating separately (Evans, 2018). Also, the finding that groups of diverse problem solvers can outperform groups of high-ability problem solvers is an argument in favor of this assumption (Hong, Page, 2004).

Of course, computations conducted in the section should be considered just an illustration of the usage of the model. Even if basic assumptions are correct, the empirical value of *I* is extremely hard to measure correctly. But if we are certain about underlying assumptions, we can conclude that singularity is very probable in the observable future (decades or centuries), independently of a specific value of *I*. Exponential character of the model implies that order-of-magnitude error in the estimation of *I* results in a linear error in the estimation of $t_s$ which is less than 5 years and is constant for every order of magnitude of *I*. It means that if we, for example, overestimate the level of development of artificial intelligence 100 000 times, the resulting error in $t_s$ will be less than 25 years. But this conclusion holds only under specific assumptions. The more general case is presented in the next section.

### 3.3. Extensions of the model

In a more general case, AI growth rate may depend on the intelligence of the engineer non-linearly:

$$\frac{dA}{A^n} = kdt \tag{13}$$

Then, growth dynamics function for *n*>1 is

$$A = \left(-\frac{1}{(-kt_2 - \frac{I^{1-n}}{1-n})(1-n)}\right)^{\frac{1}{n-1}} \tag{14}$$

Corresponding time of reaching the singularity after the moment when $A = I$:

$$t_2 = \frac{-1}{(1-n)kI^{n-1}} \tag{15}$$

Naturally, *n* must be larger than 1 (when *n=1*, the growth dynamics are exponential, hence no finite-time singularity). It is interesting to investigate limit cases for $t_2$:

$$\lim_{n \to 1} \frac{-1}{(1-n)kI^{n-1}} = \infty \tag{16}$$

$$\lim_{n \to \infty} \frac{-1}{(1-n)kI^{n-1}} = 0 \tag{17}$$

The moment of singularity is thus defined by the initial growth rate of artificial intelligence, the level of human intelligence relative to the level of artificial intelligence, and the nature of the relationship between the intellect of the engineer and the rate of growth of artificial intelligence.

But if the growth rate of intelligence depends on the intelligence of the engineer logarithmically, the singularity will never be reached:

$$dA = \ln(A)\, Akdt \tag{18}$$

Then, growth dynamics function is:

$$A = e^{e^{c+kt}} \tag{19}$$

This may be the case, as some researches argue (Moravec, 1999). Kurzweil supports this type of dynamics with empirical estimates (Kurzweil, 2010).

Hence, there are some possible positive relationships between the intellect of the engineer and the rate of growth of artificial intelligence that do not lead to singularity. Obviously, we managed to find such an example even among basic elementary functions. Of course, all functions with growth rates smaller than of the logarithmic function will also not lead to finite-time singularity.

It is, anyway, impossible to define function *I* based only on our current trend of growth. Since for every function, *I* and artificial intelligence growth rate *R* there will always be some *k* that satisfies equation (2):

$$k = \frac{lnR}{I} \tag{20}$$

Thus, other methods need to be used in order to understand the patterns of growth after $A \geq I$. These methods may rely on the assumption that the transit from exponential growth dynamics to hyperbolical growth dynamics will not be instantaneous. If artificial intelligence starts to accelerate growth rates before $A = I$, we can observe the nature of the future (probably) hyperbolical growth before $A = I$.

### 3.4. Stochastic dynamics models

Let us add a stochastic term to the model. A stochastic process is defined as:

$$dA = a_A(t, W(t))dt + b_A(t, W(t))dW(t) \tag{21}$$

We are interested in the investigation of the behavior of models described above in the case when the uncertainty is added. For this purpose, we will use the concept of time-average growth rate (Peters, 2019) which allows to track the most likely behavior of a given stochastic process (for large enough time). Time average value of X will be denoted as $X_t$.

Let us determine the time-average growth rate of the model described in subsection 3.2 with a stochastic term added:

$$dA = kIAdt + \sigma IAdW(t) \tag{22}$$

In the exponential stage, the result is well-known (Hull, 2009):

$$A_t = e^{(kI - \frac{1}{2}I\sigma^2)t_1} \tag{23}$$

But in hyperbolic stage stochastic dynamics differ. Let us assume that growth volatility is proportional to the artificial intelligence development level:

$$dA = kA^2 dt + \sigma A^2 dW(t) \tag{24}$$

Let us try to find an appropriate ergodicity transformation $u$ (Peters, 2018):

$$du = a_u dt + b_u dW(t) \tag{25}$$

where $a_u$, $b_u$ are constants. Hence, $u$ should be an arithmetic Brownian motion (for which the ergodic property (Birkhoff's equation, Moore, 2015) holds).

The following condition must hold:

$$\frac{\partial u}{dA} = \frac{b_u}{\sigma A^2} \tag{26}$$

As a result of Ito`s formula and condition (25), this must hold:

$$a_A = \frac{a_u}{b_u}\sigma A^2 + \frac{1}{2}A^2 2\sigma^2 A = A^2(\frac{a_u}{b_u}\sigma + \sigma^2 A) \tag{27}$$

We know that $a_A = kA^2$. Then,

$$a_u = (k - \sigma^2 A)\frac{b_u}{\sigma} \tag{28}$$

Generally, $a_u$ is not a constant ($a_u$ depends on $A$), and thus time-average growth dynamics cannot be defined. But, from (26) we know that:

$$u = -\frac{b_u}{A} \tag{29}$$

Hence,

$$a_u = (k + \frac{\sigma^2}{U})\frac{b_u}{\sigma} \tag{30}$$

When $U \to -\infty$, (in other words, for small enough $A$) approximate time average can be found. This is evident, since the corresponding hyperbolic function grows very slowly and almost linery for small $t$.

But in the hyperbolic stage of the growth, time average trajectory does not exist. Let us use Monte Carlo simulation to illustrate this. We approximate formula 24 with parameters $k=0.05$, $\sigma=0.05$. The results demonstrate that the singularity is being reached at significantly different points in time (from 25 to 200 periods, with 200 periods simulated), whereas in the corresponding deterministic model (formula 7) the singularity is reached in 25 periods. In some simulations, the singularity is not reached at all, and in all these cases the simulated growth dynamics appear to differ a lot.

There is no unequivocal growth trajectory, and the growth explosions (which practically can be equivalent to finite-time singularity) occur at significantly different points in time, for example, in approximately 26, 85, or 180 periods.

In some cases (for large enough $\sigma$ compared to $k$) the hyperbolic-like growth explosion is not reached for any reasonable time. Figure 3 illustrates the simulated dynamics for the parameters $k = 0.01$, $\sigma = 0.1$, and for 200 periods. Note that in the corresponding deterministic model (formula 7) the singularity is reached in 100 periods.

This can cancel the singularity even if the dynamics are hyperbolic. In other words, if we add a stochastic term to the finite-singularity model, large fluctuations may make the singularity impossible in a reasonable time.

Another important Monte-Carlo observation is that higher volatility makes hyperbolic dynamics look like exponential one (figure 5). Hence, in the case of a high volatility, the hyperbolic growth trend may be mistaken for exponential growth trend. On the other hand, the exponential growth trend cannot be mistaken for the hyperbolic one.

But time-average growth rate can be defined if convenient $a_A$ is applied, namely:

$$a_A = \frac{a_u}{b_u} b_A(A) + \frac{1}{2} b_A(A) b_A^{'}(A) \qquad (31)$$

In this case, the time-average growth rate will exist. However, whether corresponding terms $b_A$ and $a_A$ are empirically relevant, is questionable.

The same considerations can be applied to the general model described in subsection 3.3.

### 3.5. Some other ways to reach the finite-time singularity

But can singularity be achieved without resorting to a recursively improving artificial intelligence that is smarter than humans? There are some factors that can lead to singularity even if the intelligence level of the engineer is not improving. For example, if the growth of artificial intelligence and of GDP $Y$ depends correspondingly on the level of artificial intelligence and GDP:

$$\begin{cases} dY = k_1 Y A dt \\ dA = k_2 Y A dt \end{cases} \tag{31}$$

Then, if $A=1$ for $t=0$,

$$A = \frac{1}{1-k_1 t} \tag{32}$$

Instead of GDP there can be another factor similarly influencing the growth of artificial intelligence (technological progress, computational power, investment in AI, AI-corporations revenues, etc.). In general, if there is a vector $X$ of $n$ factors and level of AI development $A$ (which is a scalar), and for every element $E_i$ of the vector (A, X) holds

$$dE_i = k_i \prod_{j=1}^{n+1} E_j dt \tag{33}$$

then finite-time singularity will happen. Of course, it is also true for

$$dE_i = f(\prod_{j=1}^{n+1} E_j) dt \tag{34}$$

where $f$ is a function that $f(x)$ grows faster than $kx$.

On the other hand, hyperbolic growth can be prevented if the corresponding function grows slowly enough. The implications will be similar to the implications for the result (18). Generally, it appears to be true that effectively any reasonable connection between AI development and underlying variables represented by function *F* will not lead to finite-time singularity if the function *F* of the underlying variables grows slowly enough.

### 3.6. The reverse task

Let us solve the reverse task. What is the general view of the function, so that equation (1) is true?

Consider general continuous function *F(A)* which indicates how much AI improves depending on its current level. In other words,

$$dA = F(A)dt \tag{35}$$

$$t = \int_1^{A(t)} \frac{dA}{F(A)} \tag{36}$$

Condition (1) in terms of (36) means that

$$n = \int_1^{\infty} \frac{dA}{F(A)} \tag{37}$$

A reasonable assumption is that *F(A)* must be monotonous. So does *F(A)$^{-1}$*. From calculus, we know that the integral converges if *F(A)* has asymptotic behavior faster than *A$^{1+s}$*, where s is any positive number. Sadly, there is no boundary function with which we could compare all others and know whether the integral converges or diverges.

### 4. Discussion.

The key advantage of the model is its robustness to the assumptions. It demonstrates that if AI becomes a better AI engineer than humans, the assumptions leading to finite-time singularity are quite general. Consequently, if the finite-time singularity is achieved, it necessarily implies a significant impact on key macroeconomic and technological parameters of human civilization, contrary to infinite-time definitions of the singularity. However, high enough volatility (random deviations of the AI growth from the trend) may prevent occurring of the finite-time singularity, even if the corresponding determenistic growth model leads to it. Hence, the advantages of the model reveal themselves only in the case if stochastic effects are small enough. Practically, large intelligence explosions that happened in the simulation of stochastic hyperbolic models in subsection 3.4 can be equivalent to the effects of finite-time singularity, although in this case "infinite" level of AI development is not reached mathematically. The question which effectively needs to be answered here is: whether this explosion will be large enough so that change of the regime (phase transition) consistent with the hyperbolic growth model happens?

Empirically, the growth of AI seems to be exponential or double-exponential, which theoretically can be interpreted in various ways. In terms of the models described in the paper, this may be consistent with 3 possible scenarios:

1. The first stage of deterministic AI development with human engineers.
2. AI-self-improvement feedback loop under the assumption that AI growth rates depend on the intelligence of the engineer (or similar variable) logarithmically.
3. Hyperbolic stage of stochastic AI development with a high level of volatility.

It is impossible to make a choice among these 3 options solemnly based on the framework of the discussed models. Nevertheless, the first option is likely to be the most realistic, since AI did not exceed human intelligence yet in any general sense, and we did not observe the change of the growth regime in recent time. The third option, in principle, can be distinguished from the others, if we can observe the level of volatility consistent with it.

Whereas hyperbolic growth models developed in the study are mathematically a mere variation of the existing AI development models, they lead us to several essentially novel results and methods, namely:

1. The illustration of the generality of the assumptions leading to finite-time singularity.
2. Determination of the relative advantages of the concept of finite-time singularity.
3. Introduction of stochastic technological singularity models and investigation of the impact of stochasticity on the AI development trends.
4. A mathematical definition of the conditions leading to finite-time singularity.

Several theoretical and empirical questions related to the study remain unanswered. Further research may focus on the following problems:

- Application of the stochastic calculus to the technological singularity in general. Although we have examined the effect of stochasticity on hyperbolic growth models, its impact can be significant in the case of weak singularity models as well;
- Further empirical testing of the proposed models, including the tests that attempt to select the most empirically relevant explanation of the observed exponential or double-exponential trends and the test that determines an exact connection between the level of intelligence of the engineer and AI growth rate;
- If the proposed mechanics of hyperbolic growth regime is correct (AI growth rate depends on the level of intelligence of the engineer), a question of high importance is a timely indication of the moment when the exponential trend is substituted by the hyperbolic trend (in other words, the moment when AI exceed human intelligence and intelligence explosion begins). An establishment of a kind of a "barometer" tracking the deviations of the growth trend from exponential one may be reasonable;

- Development of other deterministic finite-time singularity models and a more broad study of empirical conditions that may lead to it. Especially, it is worth investigating which economic models can transform into finite-time singularity;
- Empirical estimation of the stochasticity of the AI development models. Developed stochastic models can be applied only if the corresponding stochastic process is Itô process (transformation of Brownian motion). Real-world dynamics may be different. For example, the application of fractional Brownian motion may be required;
- Investigation of the application of the developed models to other fields, such as, for instance, population dynamics in biology and evolution;
- Determination of exact conditions when the volatility is high enough to prevent the hyperbolic intelligence explosion.

In general, if it is assumed that the key effect of finite-time singularity is phase transition resulting from hyperbolic-like intelligence explosion and the growth dynamics may be affected by the stochasticity (which potentially prevents arbitrarily large finite-time growth), the question of whether this phase transition will be achieved cannot be answered relying only on the study of growth dynamics, *i. e.* without the study of artificial intelligence itself (which level of the development of AI is high enough to cause phase transition).

## 5. Conclusions.

Finite-time singularity can be achieved via permanent AI self-improvement if the function representing the connection between the intelligence of the engineer and the AI growth rate grows fast enough. Stochasticity, under reasonable assumption about growth dynamics in general, negatively affects time-average growth rates of AI and can even lead to canceling of the finite-time singularity. Moderate volatility can make hyperbolic dynamics look like exponential. Developed models are consistent with empirical evidence, although various interpretations are possible.

Specific singularity time estimated based on the developed model may vary a lot depending on 2 kinds of assumptions: assumptions about the difference in AI and human intelligence; assumptions about the connection between the intelligence of the engineer and the AI growth rate. If human intelligence exceeds AI by 100 times (which is assumed based on the estimation of the computational power measured in FLOPS) and the connection between the intelligence of the engineer and the AI growth rate is linear, then the time of singularity is approximately 12 years from the year 2019 (hence, around 2030). However, this estimation should rather be considered as a lower bound, for several reasons: firstly, the function representing the connection between the intelligence of the engineer and the AI growth rate may grow slower than the linear function; secondly, the difference between AI and human intelligence is likely to be larger; thirdly, even if computation power of computers will be equal to the computational power of the human brain, it will not lead automatically to the equality of these 2 types intelligence; fourthly, the impact of volatility can postpone the singularity moment. Hence, although it is feasible to answer the question "When the singularity is not possible yet?", the question "When it will occur with a high level of certainty?" requires further research.

There are some factors that can lead to singularity even if the intelligence level of the engineer is not improving. For example, if the growth of artificial intelligence and of GDP depends correspondingly on the level of artificial intelligence and GDP.

Monte Carlo simulations demonstrate that if a stochastic term is added to the finite-singularity model, large enough fluctuations may make the singularity impossible in a reasonable time. Hence, the conclusions described above will hold only in the case of low enough volatility.

**References.**

Anders Johansen and Didier Sornette. Finite-time singularity in the dynamics of the world population, economic and financial indices. Physica A, 294:465–502, 2001.

Andrey Korotayev. Compact mathematical models of world system development, and how they can help us to clarify our understanding of globalization processes. In George Modelski, Tessaleno Devezas, and William Thompson, editors, Globalization as an Evolutionary Process: Modeling Global Change, pages 133–161. Routledge, London, 2007.

Barrat, James (2013). Our Final Invention: Artificial Intelligence and the End of the Human Era, St. Martin's Press. Kindle Edition.

Brock, D. C., & Moore, G. E. (Eds.). (2006). *Understanding Moore's law: four decades of innovation*. Chemical Heritage Foundation.

Brynjolfsson, Erik and Andrew McAfee (2014). The Second Machine Age: Work, Progress, and Prosperity in a Time of Brilliant Technologies. WW Norton & Company, New York.

Chalmers, David (2010). "The singularity: a philosophical analysis". Journal of Consciousness Studies. 17 (9–10): 7–65.

Denning, P. J., & Lewis, T. G. (2017). Exponential laws of computing growth.

Easterly, W., & Levine, R. (2001). What have we learned from a decade of empirical research on growth? It's Not Factor Accumulation: Stylized Facts and Growth Models. *The world bank economic review*, *15*(2), 177-219.

Eden, Amnon H.; Moor, James H. (2012). Singularity hypotheses: A Scientific and Philosophical Assessment. Dordrecht: Springer. pp. 1–2. ISBN 9783642325601.

Eric J. Chaisson. The cosmic environment for the growth of complexity. Biosystems, 46(1-2):13 – 19, 1998.

Francois Meyer and Jacques Vallee. The dynamics of long-term growth. Technological forecasting and social change, 7:285–300, 1975.

Golosovsky, M. (2010). Hyperbolic growth of the human population of the Earth: Analysis of existing models. In *History & Mathematics: Processes and Models of Global Dynamics* (pp. 188-204).

Good, I. J. (1965). Speculations concerning the first ultraintelligent machine. *Advances in computers*, *6*(99), 31-83.

H. Saleur and D. Sornette. Complex exponents and log-periodic corrections in frustrated systems. J. Phys. I France, 6(3):327–355, mar 1996.

Hans Moravec. Simple equations for Vinge's technological singularity. http://www.frc.ri.cmu.edu/~hpm/project.archive/robot.papers/1999/singularity.html.

Hendrik Hakenes and Andreas Irmen. On the longrun evolution of technological knowledge. Economic Theory, 30:171–180, 2007.

Hong, L., & Page, S. E. (2004). Groups of diverse problem solvers can outperform groups of high-ability problem solvers. *Proceedings of the National Academy of Sciences*, *101*(46), 16385-16389.

Hull, J. (2009). Options, futures and other derivatives/John C. Hull. Upper Saddle River, NJ: Prentice Hall,.

Jones, C. I. (2016). The facts of economic growth. In *Handbook of macroeconomics* (Vol. 2, pp. 3-69). Elsevier.

Jones, Charles I. (1995). "R&D-Based Models of Economic Growth," Journal of Political Economy, 103 (4), 759-784

Jones, Charles I. (1995). "Time Series Tests of Endogenous Growth Models," Quarterly Journal of Economics, 110(2), 495-525

Kapitza, S. P. (1996). The phenomenological theory of world population growth. *Physics-uspekhi*, *39*(1), 57.

Kremer M. Population growth and technological change: One million b.c. to 1990. Quarterly Journal of Economics, 108:681–716, 1993.

Kurzweil, R. (2005). *The singularity is near: When humans transcend biology*. Penguin.

Kurzweil, R. (2010). The singularity is near. Gerald Duckworth & Co.

Lloyd, S. (2000). Ultimate physical limits to computation. Nature, 406(6799), 1047.

Luı́s M. A. Bettencourt, José Lobo, Dirk Helbing, Christian Kühnert, and Geoffrey B. West. Growth, innovation, scaling, and the pace of life in cities. Proceedings of the National Academy of Sciences, 104(17):7301–7306, April 2007.

Moore, C. C. (2015). Ergodic theorem, ergodic theory, and statistical mechanics. *Proceedings of the National Academy of Sciences*, *112*(7), 1907-1911.

Müller, V. C., & Bostrom, N. (2016). "Future progress in artificial intelligence: A survey of expert opinion". In V. C. Müller (ed): Fundamental issues of artificial intelligence (pp. 555–572). Springer, Berlin. http://philpapers.org/rec/MLLFPI

Nordhaus, W. D. (2015). Are we approaching an economic singularity? Information technology and the future of economic growth (No. w21547). National Bureau of Economic Research.

Peters, O. The ergodicity problem in economics. Nat. Phys. 15, 1216–1221 (2019). https://doi.org/10.1038/s41567-019-0732-0

Peters, O., & Adamou, A. (2018). Ergodicity Economics. *London Mathematical Laboratory*.

R Taagepera. People, skills, and resources: an interaction model for world population growth. Technological forecasting and social change, 13:13–30, 1979

Ray J. Solomonoff. The time scale of artificial intelligence: reflections on social effects. Nort-Holland Human Systems Management, 5:149–153, 1985.

Ray Kurzweil, The Singularity is Near, pp. 135–136. Penguin Group, 2005.

Ray Kurzweil, Vernor Vinge, and Hans Moravec. Singularity math trialogue. http://www.kurzweilai.net/meme/frame.html?main=/articles/art0151.html.

Robin Hanson. Economics of brain emulations. In Peter Healey and Steve Rayner, editors, Unnatural Selection - The Challenges of Engineering Tomorrow's People, pages 150–158. EarthScan, London, 2008

Robin Hanson. Economics of the singularity. IEEE Spectrum, pages 37–42, June 2008.

Robin Hanson. Is a singularity just around the corner? what it takes to get explosive economic growth. Journal of Evolution and Technology, 2, 1998. http://hanson.gmu.edu/fastgrow.html.

Roland Wagner-D¨obler. Rescher's principle of decreasing marginal returns of scientific research. Scientometrics, 50(3), 2001.

Romer, Paul M. (1986). "Increasing Returns and Long-Run Growth." Journal of Political Economy, 94 (5), 1002-37

Romer, Paul M. (1990). "Endogenous Technological Change," Journal of Political Economy. 98 (5), S71-S102.

Sandberg, A. (2016). Energetics of the brain and AI. *arXiv preprint arXiv:1602.04019*.

Sandberg, Anders. "An overview of models of technological singularity." Roadmaps to AGI and the Future of AGI Workshop, Lugano, Switzerland, March. Vol. 8. 2010.

Schmidt, E., & Cohen, J. (2013). The new digital age: Reshaping the future of people. *Nations and Business*, *66*.

Sergey P. Kapitza. Global population blow-up and after: the demographic revolution and information society. Report to the Club of Rome, 2006.

St BT Evans, J. (2018). Human versus machine thinking: the story of chess: Thinking deeply: where machine intelligence ends and human creativity begins, by Garry Kasparov, London, John Murray, 2017, 287 pp.,£ 15.42 (hardback).

Theodore Modis. Forecasting the growth of complexity and change. Technological Forecasting and Social Change, 69:377–404, 2002.

Vinge, Vernor. "The Coming Technological Singularity: How to Survive in the Post-Human Era", in Vision-21: Interdisciplinary Science and Engineering in the Era of Cyberspace, G. A. Landis, ed., NASA Publication CP-10129, pp. 11–22, 1993.

Yampolskiy, Roman V. "Analysis of types of self-improving software." Artificial General Intelligence. Springer International Publishing, 2015. 384-393.

**Figures Legends**

*Figure 1. Example of exponential growth dynamics in linear and logarithmic (log) scales*

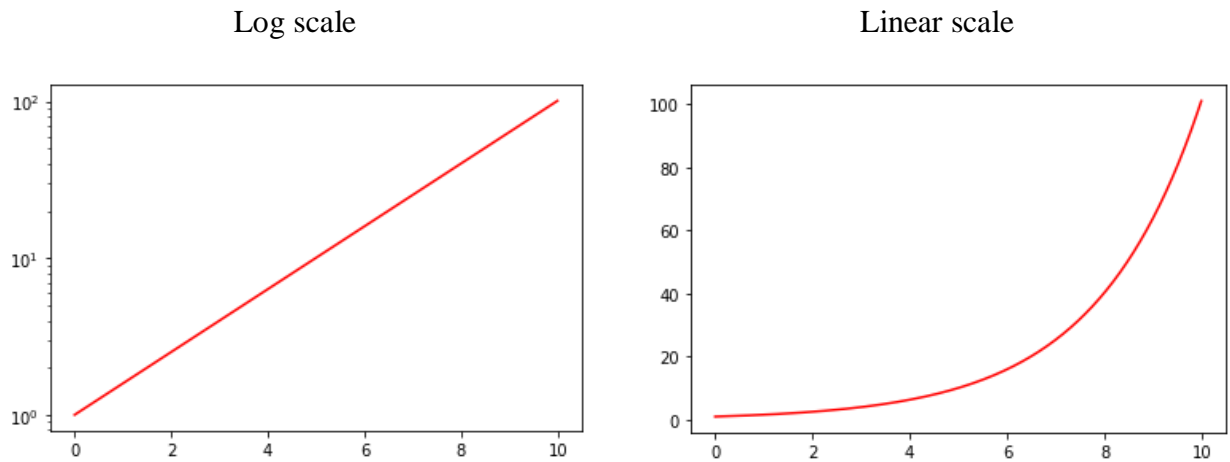

Source: own results. Created via Python Matplotlib

*Figure 2. Example of hyperbolic growth dynamics in linear and logarithmic (log) scales*

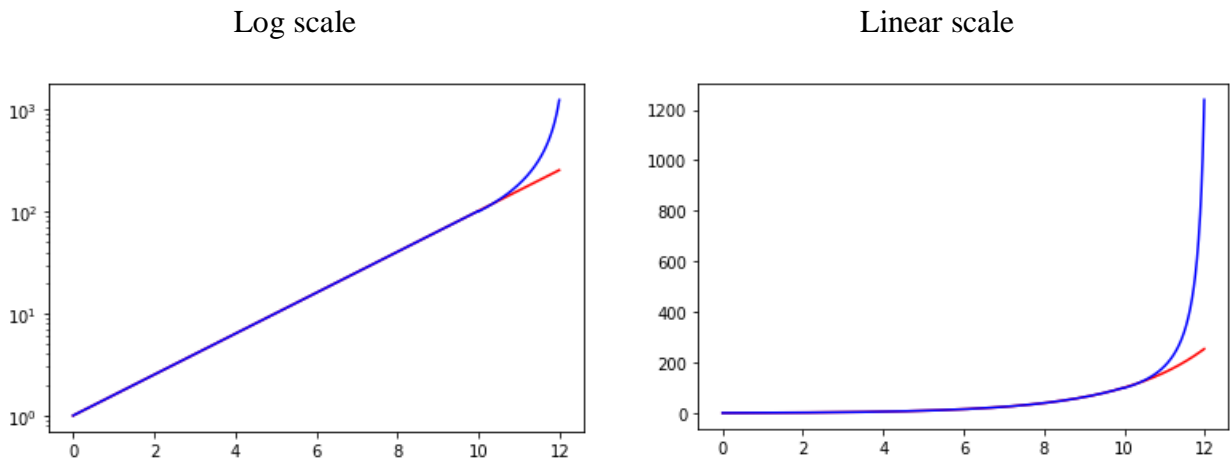

Source: own results. Created via Python Matplotlib

*Figure 3. Simulation of stochastic hyperbolic growth dynamics (formula 24) for parameters k=0.05, σ=0.05 in the cases when the finite-time singularity is not reached in 200 periods*

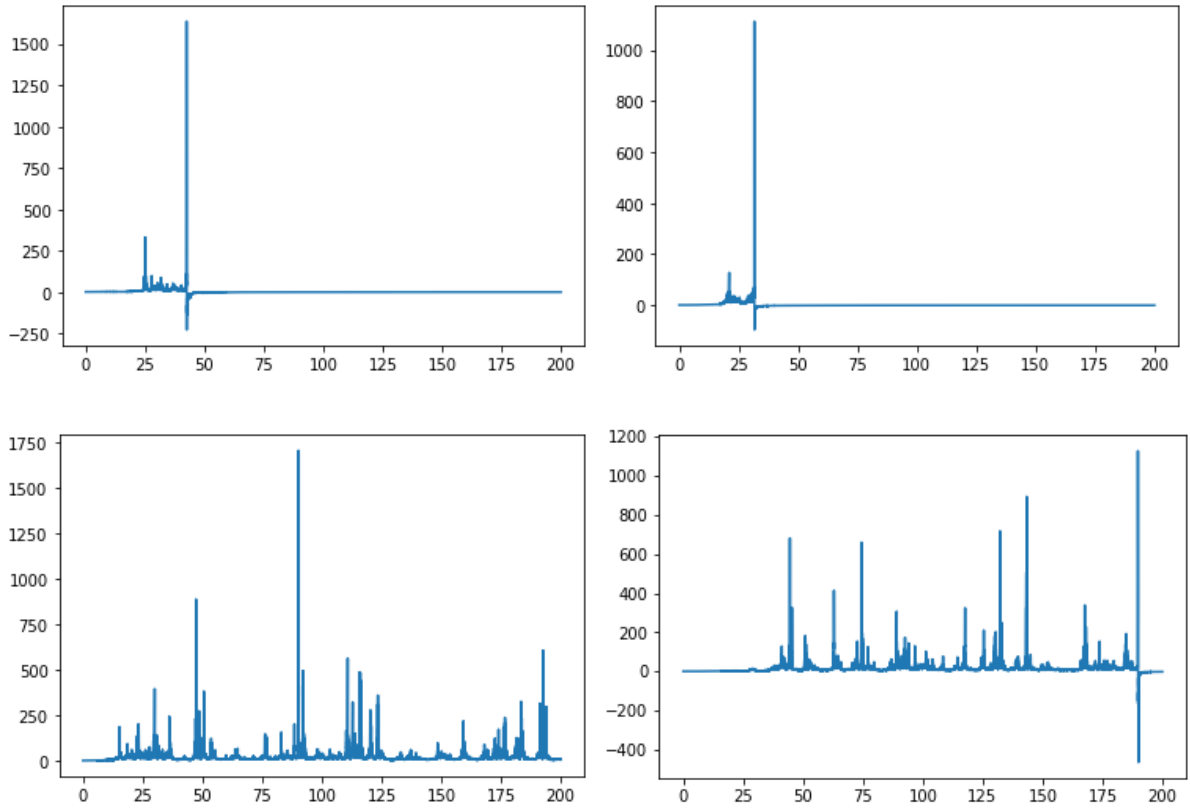

Source: own results. Created via Python Matplotlib

*Figure 4. Simulation of stochastic hyperbolic growth dynamics (formula 24) for parameters k=0.01, σ=0.1*

Source: own results. Created via Python Matplotlib

*Figure 5. Simulation of stochastic hyperbolic growth dynamics (formula 24) for parameters k=0.01, σ=0.001, 0.02*

σ=0.001              σ=0.02

Source: own results. Created via Python Matplotlib